\documentclass[letterpaper]{article} % DO NOT CHANGE THIS
\usepackage{aaai25}  % DO NOT CHANGE THIS
\usepackage{times}  % DO NOT CHANGE THIS
\usepackage{helvet}  % DO NOT CHANGE THIS
\usepackage{courier}  % DO NOT CHANGE THIS
\usepackage[hyphens]{url}  % DO NOT CHANGE THIS
\usepackage{graphicx} % DO NOT CHANGE THIS
\usepackage{natbib}  % DO NOT CHANGE THIS AND DO NOT ADD ANY OPTIONS TO IT
\usepackage{caption} % DO NOT CHANGE THIS AND DO NOT ADD ANY OPTIONS TO IT
\usepackage{algorithm}
\usepackage{algorithmic}

\usepackage{color}
\usepackage{url}
\usepackage{graphicx} 
\usepackage{adjustbox} 
\usepackage{amsfonts}
\usepackage{amsmath}
\usepackage{amssymb}
\usepackage{bm}
\usepackage{xcolor}
\usepackage{multirow}
\usepackage{booktabs}
\usepackage{graphicx}
\usepackage{newfloat}
\usepackage{listings}
\DeclareCaptionStyle{ruled}{labelfont=normalfont,labelsep=colon,strut=off} % DO NOT CHANGE THIS
\floatstyle{ruled}
\newfloat{listing}{tb}{lst}{}
\floatname{listing}{Listing}
\title{DocMamba: Efficient Document Pre-training with State Space Model}
\author{
	Pengfei Hu\textsuperscript{\rm 1\equalcontrib },
	Zhenrong Zhang\textsuperscript{\rm 1, 2\equalcontrib \footnote{Work done during an internship at iFLYTEK Research.} },
	Jiefeng Ma\textsuperscript{\rm 1},
	Shuhang Liu\textsuperscript{\rm 1},
	Jun Du\textsuperscript{\rm 1}\thanks{Corresponding author.},
	Jianshu Zhang\textsuperscript{\rm 2}
}

\affiliations{
	\textsuperscript{\rm 1}NERC-SLIP, University of Science and Technology of China\\
	\textsuperscript{\rm 2}iFLYTEK Research\\
	\{pengfeihu,zzr666,jfma,liush19\}@mail.ustc.edu.cn, jundu@ustc.edu.cn, jszhang6@iflytek.com
}

\usepackage{bibentry}
\begin{document}

\maketitle

\begin{abstract}
In recent years, visually-rich document understanding has attracted increasing attention. Transformer-based pre-trained models have become the mainstream approach, yielding significant performance gains in this field. However, the self-attention mechanism's quadratic computational complexity hinders their efficiency and ability to process long documents. In this paper, we present DocMamba, a novel framework based on the state space model. It is designed to reduce computational complexity to linear while preserving global modeling capabilities. To further enhance its effectiveness in document processing, we introduce the Segment-First Bidirectional Scan (SFBS) to capture contiguous semantic information. Experimental results demonstrate that DocMamba achieves new state-of-the-art results on downstream datasets such as FUNSD, CORD, and SORIE, while significantly improving speed and reducing memory usage. Notably, experiments on the HRDoc confirm DocMamba's potential for length extrapolation. 
\end{abstract}

 \begin{links}
     \link{Code}{https://github.com/Pengfei-Hu/DocMamba}
 \end{links}

\section{Introduction}
With the prosperity of commercial activities in today’s society,  a broad range of documents are used to convey information, leading to a growing demand for document processing \cite{documentai}. In order to reduce the labor-intensive workflows associated with this, Visually-rich Document Understanding (VrDU) \cite{layoutlm} is drawing considerable attention from both academia and industry. It aims to automate information extraction from documents \cite{semv2,cdc} and support various applications.

In recent years, Transformer-based \cite{transformer} pre-training models have made substantial advancements in VrDU and become the mainstream practice. The pioneering model, LayoutLM \cite{layoutlm}, encodes both textual and layout information through an architecture similar to BERT \cite{bert}. Subsequent research \cite{selfdoc,layoutlmv2,layoutlmv3} has incorporated additional visual models as image encoders, enabling the joint modeling of visual, textual, and layout information. However, despite the significant boost provided by Transformer, its quadratic complexity with respect to input length limits its ability to handle long texts. For instance, the context size in the LayoutLM series \cite{layoutlm,layoutlmv2,layoutlmv3} is restricted to 512. Consequently, when processing text-dense documents, these models often need to incorporate a sliding window strategy, which can lead to the loss of global information and an increase in processing time.

\begin{figure}[t]
	\centering
	\includegraphics[width=0.95\columnwidth]{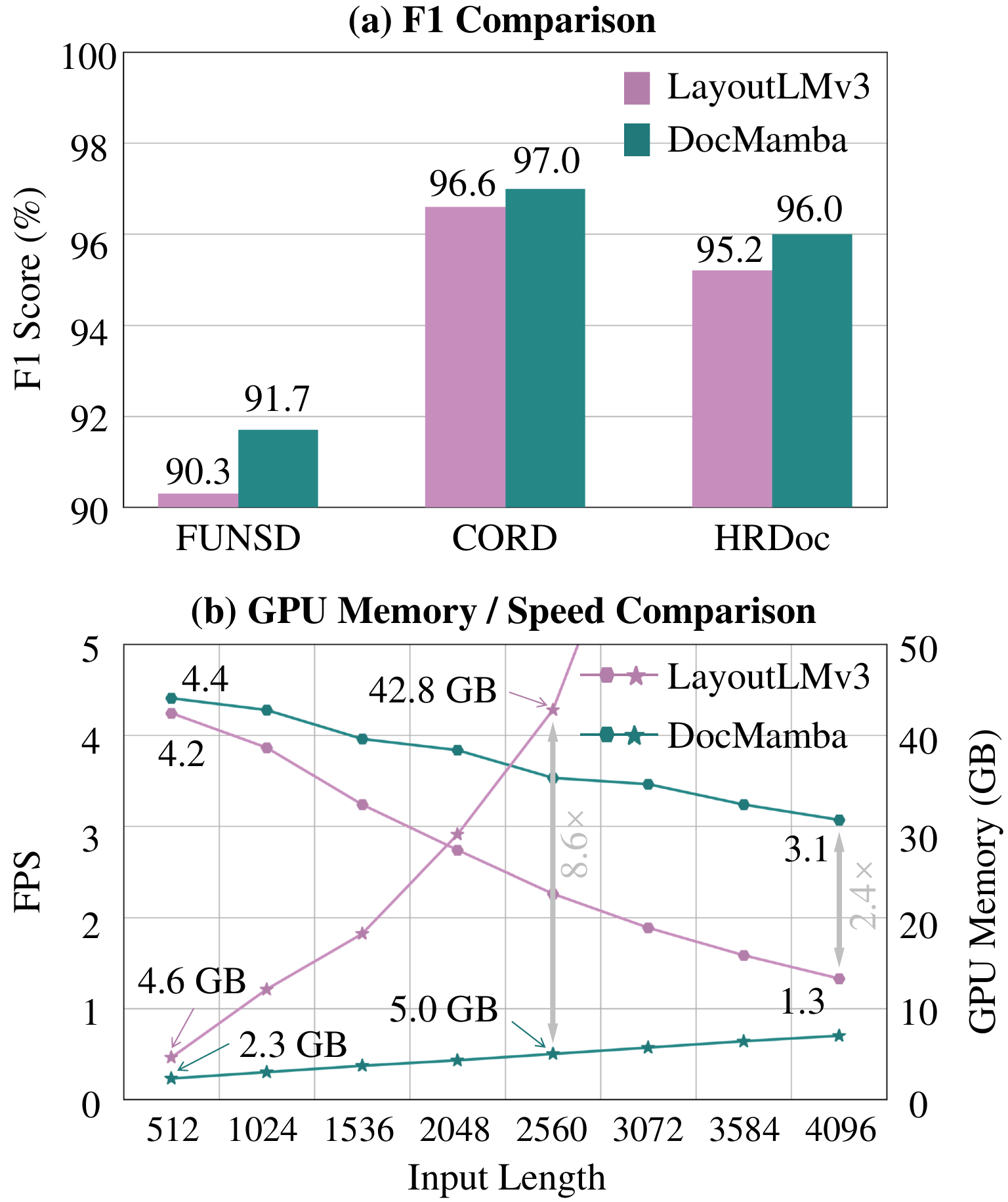} % Reduce the figure size so that it is slightly narrower than the column. Don't use precise values for figure width.This setup will avoid overfull boxes.
	\caption{Performance and efficiency comparisons between LayoutLMv3 \cite{layoutlmv3} and our DocMamba.}
	
	\label{img:simple_result}
\end{figure}

To achieve sub-quadratic complexity, one promising approach is the substitution of the Transformer with State Space Models (SSMs) \cite{s4d}. They originated from the foundational classic state space model \cite{ssm}, and are notable for their capabilities in linear-time inference, highly parallelized training, and robust performance in tasks requiring long-context processing. Examples include the linear state space layers (LSSL) \cite{lssl} and the structured state-space sequence model (S4) \cite{s4}. A recent addition to this category, Mamba \cite{mamba}, has demonstrated exceptional results through its selective mechanism and hardware-aware design. Unlike self-attention mechanism in that each token interacts with all others within the context, Mamba enables each token to garner contextual knowledge solely through a compressed hidden state, thereby reducing the quadratic complexity to linear. Mamba has shown performance comparable to the Transformer in various fields \cite{jamba,visionmamba}. Given the inherently longer sequences produced by documents, a natural question arises: Can Mamba work well for VrDU?

Motivated by this, we introduce DocMamba, a purely SSM-based model tailored for VrDU. It boasts linear complexity relative to input length, making it ideal for long documents. While vanilla Mamba is designed to process 1-D sequences, tokens in documents exhibit complex 2-D layouts and form continuous semantic content alongside their neighbors. Thus, it is necessary to consecutively process tokens belonging to the same segment (e.g., titles, paragraphs, captions). For this purpose, we design the Segment-First Bidirectional Scan (SFBS). Initially,  we leverage existing document layout analysis systems \cite{publaynet} to extract segments. DocMamba then sequentially scans all tokens within one segment before shifting to the next. Considering that incorporating context from both directions enhances the performance of language models \cite{bert}, we adopt the bidirectional scan strategy following Vim \cite{visionmamba}. Furthermore, due to the inherent positional information within SSMs, DocMamba does not require 1-D position embeddings, which are indispensable in Transformer-based models. This feature endows DocMamba with the potential for length extrapolation.

We evaluate the performance of the pre-trained DocMamba using several publicly available benchmark datasets in downstream tasks. As depicted in Figure \ref{img:simple_result} (a), DocMamba surpasses the strong baseline LayoutLMv3 \cite{layoutlmv3} at the base scale with a similar number of parameters across three datasets: the FUNSD dataset \cite{funsd} for form understanding, the CORD \cite{cord} dataset for receipt understanding, and the HRDoc \cite{hrdoc} for semantic unit classification. Moreover, as shown in Figure \ref{img:simple_result} (b), tests on the HRDoc dataset show DocMamba has a faster inference speed and less GPU memory usage than LayoutLMv3. Especially with larger input lengths, DocMamba can save up to 88.3\% of GPU memory and work 2.4 times faster, which can reduce the application costs significantly. This also proves the linear computational complexity of DocMamba. Furthermore, when the input length is restricted to 512 during both pre-training and fine-tuning, DocMamba still yields impressive results when the token length of test samples reaches 2560 for semantic unit classification on the HRDoc dataset. This validates DocMamba's potential in length extrapolation. In conclusion, our research underscores the potential of SSMs as a powerful competitor with Transformer for VrDU, offering a simple yet effective baseline for future research.

Our main contributions are listed as follows:
\begin{itemize} 
	\item We delve into the SSM-based VrDU and propose a novel method, DocMamba, which exhibits linear complexity with respect to input length. 
	\item We introduce the Segment-First Bidirectional Scan (SFBS) to enable Mamba, initially designed for 1-D sequences, to effectively process document tokens that possess complex 2-D layouts. 
	\item Extensive experiments demonstrate that DocMamba exhibits promising performance compared to strong Transformer-based models, while maintaining faster speeds, lower memory consumption, and the potential for length extrapolation. 
\end{itemize}

\begin{figure*}[t]
	\centering
	\includegraphics[width=1.85\columnwidth]{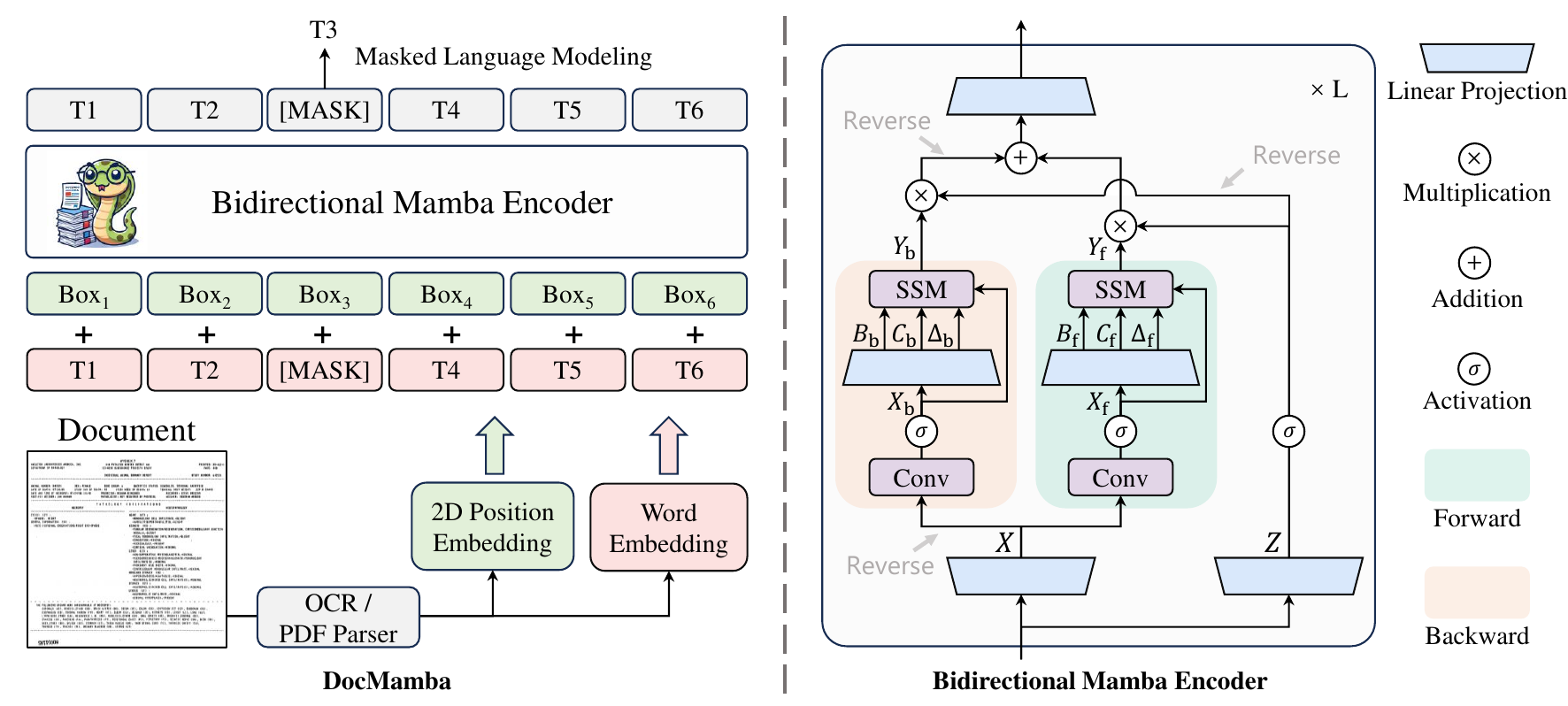} % Reduce the figure size so that it is slightly narrower than the column. Don't use precise values for figure width.This setup will avoid overfull boxes.
	\caption{Framework of DocMamba (left) and Bidirectional Mamba Encoder (right).}
	\label{img:pipeline}
\end{figure*}

\begin{figure}[t]
	\centering
	\includegraphics[width=0.85\columnwidth]{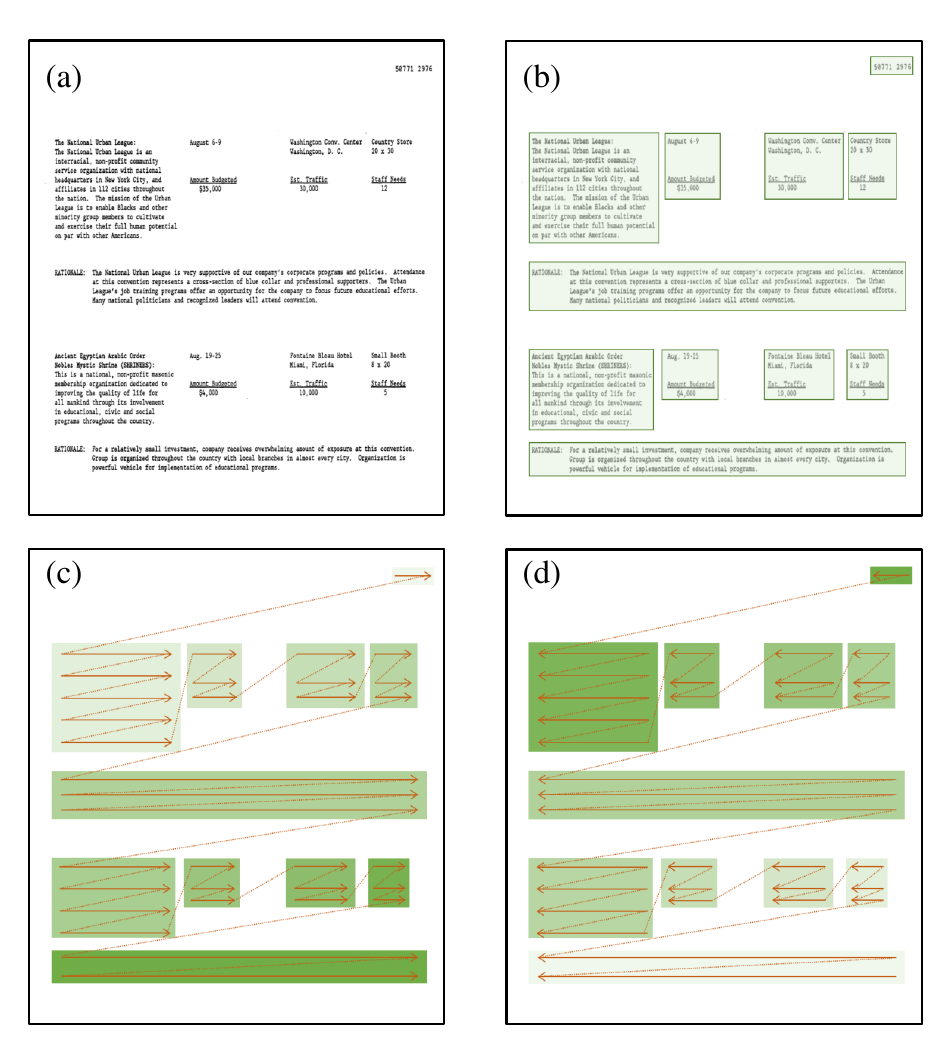} % Reduce the figure size so that it is slightly narrower than the column. Don't use precise values for figure width.This setup will avoid overfull boxes.
	\caption{Depiction of Segment-First Bidirectional Scan.}

	\label{img:sfbs}
\end{figure}

\vspace*{-10pt}

\section{Related Work}

\vspace*{-2pt}

\subsection{Visually-rich Document Understanding}

Early research \cite{yang2017learning,mtd} in VrDU typically utilizes unimodal or multimodal models with shallow fusion techniques. In recent years, the advent of pre-training techniques has revolutionized this field. BERT \cite{bert} uses masked language models to obtain pre-trained deep bidirectional representations within pure text. Inspired by BERT, LayoutLM \cite{layoutlm} introduces 2-D spatial coordinate embeddings in addition to 1-D positional and text embeddings, thus simultaneously modeling the interaction between text and layout information within a singular framework. Furthermore, LayoutLMv2 \cite{layoutlmv2} adapts the standard Transformer by integrating a spatial-aware self-attention mechanism, and concatenates visual tokens with textual tokens to enhance text-image interactions. LayoutLMv3 \cite{layoutlmv3} suggests learning cross-modal alignment with unified text and image masking. Additionally, various model architectures \cite{docformer,xylayoutlm}, attention mechanisms \cite{bros,graphdoc} and self-supervised
tasks \cite{layoutmask,geolayoutlm,unitabnet,yao2024swift} have been explored. However, nearly all of these methods are based on Transformer, which has a quadratic complexity concerning input length, thus posing challenges when processing lengthy documents.

\vspace*{-5pt}

\subsection{State Space Models}
State Space Models (SSMs) serve as a fundamental model applied across various fields such as control theory \cite{controltheory}, signal processing \cite{signalprocessing}, and applied economics \cite{economics}. Recently, SSMs have garnered renewed attention within the deep learning community \cite{lssl,s4,s5}, demonstrating notable proficiency in capturing long-range dependencies. They afford highly efficient computation, either as a recurrence or convolution operation, with linear or near-linear scalability in sequence length. Mamba \cite{mamba}, in particular, distinguishes itself by incorporating a time-varying selection mechanism and a hardware-aware parallel algorithm. The significant potential demonstrated by Mamba has inspired a succession of studies in areas like NLP \cite{jamba,crossmamba}, video understanding \cite{videomamba,videomambapro,yao2024qe}, speech processing \cite{speechmamba}, and more. However, the application of SSMs for VrDU still remains unexplored.

\vspace*{-5pt}

\section{Preliminaries}

\textbf{State Space Model.} The classical SSM represents a continuous system that maps an input $x(t) \in \mathbb{R}$ to an output $y(t) \in \mathbb{R}$ through an implicit latent state $\boldsymbol{h}(t) \in \mathbb{R}^N$. This can be typically formulated as follows:
$$
\begin{gathered}
	\boldsymbol{h}^{\prime}(t) = \boldsymbol{A} \boldsymbol{h}(t)+\boldsymbol{B} x(t) \\
	y(t) = \boldsymbol{C} \boldsymbol{h}(t)
\end{gathered}
$$

Here, $\boldsymbol{A} \in \mathbb{R} ^{N \times N}$ denotes the evolution matrix, while $\boldsymbol{B} \in \mathbb{R} ^{N \times 1}$ and $\boldsymbol{C} \in \mathbb{R} ^{1 \times N}$ denote  the input and output mapping matrices, respectively. 

\vspace{2mm}

\noindent \textbf{Discrete SSM.} For integration into deep learning models, SSM requires discretization. Specifically, $\boldsymbol{A}$, $\boldsymbol{B}$ are transformed into their discretized counterparts $\overline{\boldsymbol{A}}$, $\overline{\boldsymbol{B}}$ using a timescale parameter $\Delta \in \mathbb{R}$ \cite{s4}. This transformation commonly utilizes the Zero-Order Hold
(ZOH) method, defined by:
$$
\begin{gathered}
\overline{\boldsymbol{A}} = \exp (\Delta \boldsymbol{A}) \\
\overline{\boldsymbol{B}} = (\Delta \boldsymbol{A})^{-1}(\exp (\Delta \boldsymbol{A})-\boldsymbol{I}) \cdot \Delta \boldsymbol{B}
\end{gathered}
$$
This allows the discrete SSM to be represented as:
$$
\begin{gathered}
	\boldsymbol{h}_t =\overline{\boldsymbol{A}} \boldsymbol{h}_{t-1}+\overline{\boldsymbol{B}} x_t \\
	y_t =\boldsymbol{C} \boldsymbol{h}_t
\end{gathered}
$$

\noindent \textbf{Mamba.} As evident from the above, the parameters within SSM remain invariant with respect to the input. Mamba \cite{mamba} identifies this as a fundamental limitation of SSM. In response, Mamba introduces a selection mechanism by setting $\boldsymbol{B}$, $\boldsymbol{C}$ and $\Delta$ as functions of $x_t$, which allows for propagating or forgetting information throughout the sequence depending on the current token. Additionally, to ensure GPU efficiency, Mamba employs a hardware-aware algorithm within the selective SSM.

\section{Method}
This section delineates the core components of our DocMamba, as depicted in Figure \ref{img:pipeline}. Initially, we introduce the Segment-First Bidirectional Scan to enhance DocMamba's ability to understand tokens in documents. Following this, we introduce the model architecture in detail. The final part illustrates the pre-training of DocMamba, including the training objective and several effective training strategies.

\subsection{Segment-First Bidirectional Scan}
Vanilla Mamba, which is well-suited for the 1-D sequences, captures long-range dependencies by updating the hidden state based on the current token at each step. However, tokens in documents exhibit complex 2-D spatial layouts and share continuous semantic information in conjunction with their neighbors. Therefore, we design the Segment-First Bidirectional Scan (SFBS) to derive 1-D token sequences from documents, as demonstrated in Figure \ref{img:sfbs}.

Specifically, given a document image as illustrated in Figure \ref{img:sfbs} (a), an off-the-shelf document layout analysis system \cite{publaynet} is first employed to extract segments such as titles, paragraphs, and captions as depicted in Figure \ref{img:sfbs} (b). The tokens within each segment are then separately arranged in an order that primarily descends along the Y-axis and then the X-axis. The order of scanning the segments follows a similar pattern. Furthermore, a bidirectional scanning strategy is adopted, as it enables each token in the document to gain global information. The final scanning orders are demonstrated in Figure \ref{img:sfbs} (c) and (d), where the lighter regions mark the initiation of SFBS, and the darker regions denote its termination.
 
\subsection{Model Architecture}
DocMamba employs a multi-layer bidirectional Mamba structure as the backbone, taking text and layout information as input. Document images are preprocessed using PaddleOCR\footnote{https://github.com/PaddlePaddle/PaddleOCR} to attain the words and corresponding 2-D positions. Detailed descriptions are as follows.

%\vspace{3mm}

\noindent \textbf{Word Embedding.} The text content is tokenized using Byte-Pair Encoding (BPE)\cite{bpe}. Each sequence always begins with a specific classification token ([CLS]). Unlike Transformer-based models that necessitate the addition of a 1-D positional embedding to denote word order within a sentence, DocMamba disregards 1-D positional embedding due to the inherent nature of the sequential order within SSMs. Therefore, the $i$-th word embedding can be formulated as:
$$
\boldsymbol{t}_i = \text{TokenEmb} (w_i)
$$
%\vspace{3mm}

\noindent \textbf{2-D Position Embedding.} Given the significant influence of a word's spatial location within a document on its semantic representation, 2-D positional embedding is employed to model these relative spatial positions. Following standard practice \cite{graphdoc,layoutlmv3,mtd}, a document page is considered a coordinate system originating at the top-left. All coordinates are normalized and discretized to integers within the range [0, 1000]. The normalized coordinate of the $i$-th text token's four vertices is denoted as $\text{poly}_i = (x_1, y_1, x_2, y_2, x_3, y_3, x_4, y_4)$, proceeding clockwise from the upper left corner. For the $t$-th element in $\text{poly}_i$, its embedding can be obtained by: 
$$ \boldsymbol{e}_{i, t} = \text{PosEmb2D}_{\text{xy}}(\text{poly}_{i,t}) + \text{CoordTypeEmb}(t)$$ 
where $\text{PosEmb2D}_{\text{xy}}$ is shared between X-axis and Y-axis, and CoordTypeEmb represents the type embedding associated with each coordinate in $\text{poly}_i$. The $i$-th 2-D position embedding is the concatenation of $\boldsymbol{e}_{i, 1} \sim \boldsymbol{e}_{i, 8}$:
$$
\boldsymbol{l}_i = \text{Concat} [ \boldsymbol{e}_{i, t} ],\ t=1,\dots,8
$$

%\vspace{3mm}

\noindent \textbf{Bidirectional Mamba Encoder.} The input embeddings $\boldsymbol{S}^0=\{\boldsymbol{s}_1^0, \boldsymbol{s}_2^0 \dots \boldsymbol{s}_N^0\}$ are computed by summing the word and 2-D position embeddings:
$$
\boldsymbol{s}_i^0 = \boldsymbol{t}_i + \boldsymbol{l}_i
$$

These input embeddings are then processed through multi-layer bidirectional Mamba blocks. Specifically, the output from the previous layer, $\boldsymbol{S}^{m-1}$, is fed into the $m$-th layer, getting the output $\boldsymbol{S}^{m}$ with a residual connection:
$$
\boldsymbol{S}^{m} = \text{BiMambaBlock}(\boldsymbol{S}^{m-1}) + \boldsymbol{S}^{m-1}
$$

BiMambaBlock denotes the bidirectional Mamba block as illustrated on the right part of Figure \ref{img:pipeline}. For the $m$-th layer, the input $\boldsymbol{S}^{m-1}$ is first normalized and linearly projected to $\boldsymbol{X}$ and $\boldsymbol{Z}$. $\boldsymbol{X}$ is subsequently processed in both forward and backward directions. In the forward process, $\boldsymbol{X}$ passes through a 1-D convolution layer followed by an activation function to produce $\boldsymbol{X}_\text{f}$. $\boldsymbol{X}_\text{f}$ is then linearly projected to generate the $\boldsymbol{B}_\text{f}$, $\boldsymbol{C}_\text{f}$, and $\boldsymbol{\Delta}_\text{f}$. These components, along with $\boldsymbol{X}_\text{f}$, are fed into the SSM to compute the discrete $\overline{\boldsymbol{A}}_\text{f}$ and $\overline{\boldsymbol{B}}_\text{f}$, leading to the SSM's output $\boldsymbol{Y}_\text{f}$. The backward output, $\boldsymbol{Y}_\text{b}$, is similarly produced by reversing $\boldsymbol{X}$ from $[\boldsymbol{x}_1;\boldsymbol{x}_2;...;\boldsymbol{x}_N]$ to $[\boldsymbol{x}_N;\boldsymbol{x}_{N-1};...;\boldsymbol{x}_1]$. The parameters for the forward and backward directions are not shared. Finally, $\boldsymbol{Y}_\text{f}$ and $\boldsymbol{Y}_\text{b}$ are gated by $\boldsymbol{Z}$ and summed to produce the output of the current block through a linear layer.

\subsection{Pre-training Strategy}
Following standard procedure \cite{layoutlm,graphdoc}, we employ Masked Language Modeling (MLM) as the pre-training task. This task enables the learning of language representation incorporating layout embedding cues. In the pre-training phase, each token is independently and randomly masked with a given probability $\text{P}_\text{mask}$, while the associated layout information remains intact. Masked tokens are replaced with a special symbol [MASK]. The output representations of the masked tokens from the encoder are fed into a classifier over the entire vocabulary.

Contrary to prior Transformer-based models that maintain a constant batch size and input length during pre-training, DocMamba is capable of dynamically adjusting the batch size based on the input length. Specifically, we allocate the sequences into non-overlapping buckets based on their lengths, with each bucket covering a range of 64. Within each bucket, input sequences are truncated to the same size. Given the input sequence of length $l$, we assign the batch size $b$ through $b = \text{k} / l$, where $\text{k}$ is a constant. This formula is effective because of the linear GPU memory consumption of DocMamba. This approach enhances the efficiency of the pre-training process and empowers the model to dynamically handle document contents of varying lengths.

\begin{table*}[t]
	\centering
\resizebox{1.82\columnwidth}{!}{
	\begin{tabular}{lclccc}
		\toprule
		\bf Model & \bf Parameters & \bf Modality & \bf FUNSD (F1$\uparrow$) & \bf CORD (F1$\uparrow$) & \bf SROIE (F1$\uparrow$)  \\
		\midrule
		$\text{BERT}$ \cite{bert} & 110M & T & 60.3 & 89.7 &  91.0 \\
		$\text{RoBERTa}$ \cite{roberta} & 125M & T & 66.5 & 93.5 &  - \\
		$\text{Mamba}$ \cite{mamba} & 130M & T & 47.5 & 74.3 &  77.0 \\
		$\text{Mamba} ^\ast$ \cite{mamba} & 130M & T & 58.3 & 85.3 &  83.2 \\
		$\text{LayoutLM}$ \cite{layoutlm} & 160M & T+L & 79.3 & - & 94.4 \\
		$\text{BROS}$ \cite{bros} & 110M & T+L & 83.1 & 95.7 &  95.5 \\
		$\text{LiLT}$ \cite{lilt} & - & T+L & 88.4 & 96.1 & - \\
		$\text{SelfDoc}$ \cite{selfdoc} & - & T+L+I & 83.4 & - & - \\
		$\text{UDoc}$ \cite{unidoc} & 272M & T+L+I & 87.9 & \textcolor{gray}{98.9}$^\dagger$  & - \\
		$\text{TILT}$ \cite{tilt} & 230M & T+L+I & - & 95.1 & \textcolor{gray}{97.7}$^\ddagger$ \\
		$\text{DocFormer}$ \cite{docformer} & 183M & T+L+I & 83.3 & 96.3 & - \\
		$\text{XYLayoutLM}$ \cite{xylayoutlm} & - & T+L+I & 83.4 & - & - \\
		$\text{LayoutLMv2}$ \cite{layoutlmv2} & 200M & T+L+I & 82.8 & 95.0 & \underline{96.3} \\
		$\text{LayoutLMv3}$ \cite{layoutlmv3} & 133M & T+L+I & \underline{90.3} & \underline{96.6} & - \\
		\midrule
		$\textbf{DocMamba}$ & 135M & T+L & \textbf{91.7} & \textbf{97.0} & \textbf{96.8} \\
		\bottomrule
		\multicolumn{6}{l}{\footnotesize
		}

	\end{tabular}
}
	\caption{Comparison with existing methods. ``T/L/I" stands for ``text/layout/image" modality. $^\dagger$: UDoc split the CORD into 626/247 receipts for training/test, deviating from the official 800/100 split for training/test, so the score is not directly comparable. $^\ddagger$: TILT employed extra supervised data for pre-training, making its score not directly comparable as well. $^\ast$: To keep a fair comparison with DocMamba, we use the same data as DocMamba to pretrain the vanilla Mamba from scratch.
	}
\label{tab:sota}
\end{table*}

\section{Experiments}

\subsection{Datasets}
We select several datasets to evaluate the performance of DocMamba, including FUNSD \cite{funsd}, CORD \cite{cord}, SROIE \cite{sroie} and HRDoc \cite{hrdoc}.

\noindent \textbf{FUNSD.} The FUNSD dataset is a noisy scanned document dataset for form understanding, containing 149 training samples and 50 testing samples. It defines the entity extraction task aimed at extracting values for predefined keys: ``question", ``answer", ``header" or ``other".

\noindent \textbf{CORD.}  The CORD dataset is used for key information extraction from receipts, comprising 800 training samples, 100 validation samples, and 100 test samples. It includes 30 semantic labels under 4 categories: ``company", ``date", ``address", and ``total".

\noindent \textbf{SROIE.} The SROIE dataset is another receipt understanding dataset, consisting of 626 training receipts and 347 test receipts. The task is the same as CORD.

\noindent \textbf{HRDoc.}  The HRDoc dataset is designed for the hierarchical reconstruction of academic document structures. We use the HRDoc-Hard subset, which includes 1,000 training documents and 500 testing documents. Our focus is on semantic unit classification, aiming to categorize each unit into one of 14 categories: ``title'', ``author'', ``mail'', ``affiliation'', ``section'', ``first-line'', ``para-line'', ``equation'', ``table'', ``figure'', ``caption'', ``page-footer'', ``page-header'', and ``footnote''. HRDoc contains text-dense documents, and we use it to validate DocMamba's potential for length extrapolation.

\subsection{Implementation Details}
DocMamba employs a 24-layer bidirectional Mamba encoder with a hidden size of 768 and an intermediate size of 1,536. For the SSM within each layer, we use the default hyperparameters from Mamba \cite{mamba}, setting the state dimension to 16. The coordinates of [CLS] are zeros.

\noindent \textbf{Pre-training.} We use 10 million pages from the IIT-CDIP Test Collection 1.0 \cite{cdip}, a large-scale scanned document image dataset, to pre-train DocMamba. The constant $\text{k}$ for computing the varying batch size of a single GPU is 20,480. For example, the batch size is set to 40 for an input length of 512. For the MLM task, following the settings in BERT \cite{bert}, we randomly mask 15\% of all input tokens. Out of these, 80\% are replaced by [MASK], 10\% are replaced by random tokens from the vocabulary, and 10\% remain unchanged. We adopt distributed training and mixed-precision training to reduce memory costs and speed up training procedures. DocMamba is pre-trained using the Adam optimizer \cite{adam} with a learning rate of $5 \times 10^{-5}$ for $500,000$ steps. The learning rate is warmed up over the first 10\% steps and then linearly decayed. Pre-training is conducted on 8 Telsa A40 48GB GPUs.

\noindent \textbf{Finu-tuning.} We treat FUNSD, CORD, and SROIE as sequential labeling tasks, using BIO tags for each entity field. We use the officially-provided images and OCR annotations and build a dropout layer and a linear layer above the output representations. DocMamba is fine-tuned on these datasets for 1,000 steps with a learning rate $2 \times 10^{-5}$ and a batch size of $16$. For HRDoc, we directly predict the categories for each unit, using a learning rate of $2 \times 10^{-5}$, a batch size of $48$ for 2,000 steps.

\subsection{Comparison With State-of-the-Art Methods}

\noindent \textbf{Comparison of F1 scores.} Table \ref{tab:sota} illustrates the performance of various methods in form and receipt understanding. These methods can be categorized by the modalities used in pre-training. ``T" represents pure text models like BERT \cite{bert} and RoBERTa \cite{roberta}. ``T+L" means text and layout models such as LayoutLM \cite{layoutlm} and BROS \cite{bros}. ``T+L+I" denotes models that incorporate text, layout, and image modalities, including LayoutLMv2 \cite{layoutlmv2}, SelfDoc \cite{selfdoc}, and LayoutLMv3 \cite{layoutlmv3}. Some methods offer different versions, such as $\rm{base}$ and $\rm{large}$, due to variations in parameter sizes. To ensure a fair comparison, we opt for the $\rm{base}$ versions of previous methods, as they maintain a similar number of parameters to that of DocMamba. The entity-level F1 score serves as our evaluation metric. Despite the absence of an image modality in DocMamba, it still outperforms all other methods, including the ``T+L+I" models across all three datasets (FUNSD by + 1.4\%, CORD by + 0.4\%, SROIE by + 0.5\%). These results attest to DocMamba's competitive performance against Transformer-based models, underscoring the substantial potential of SSMs in VrDU.

\begin{figure}[t]
	\centering
	\includegraphics[width=0.99\columnwidth]{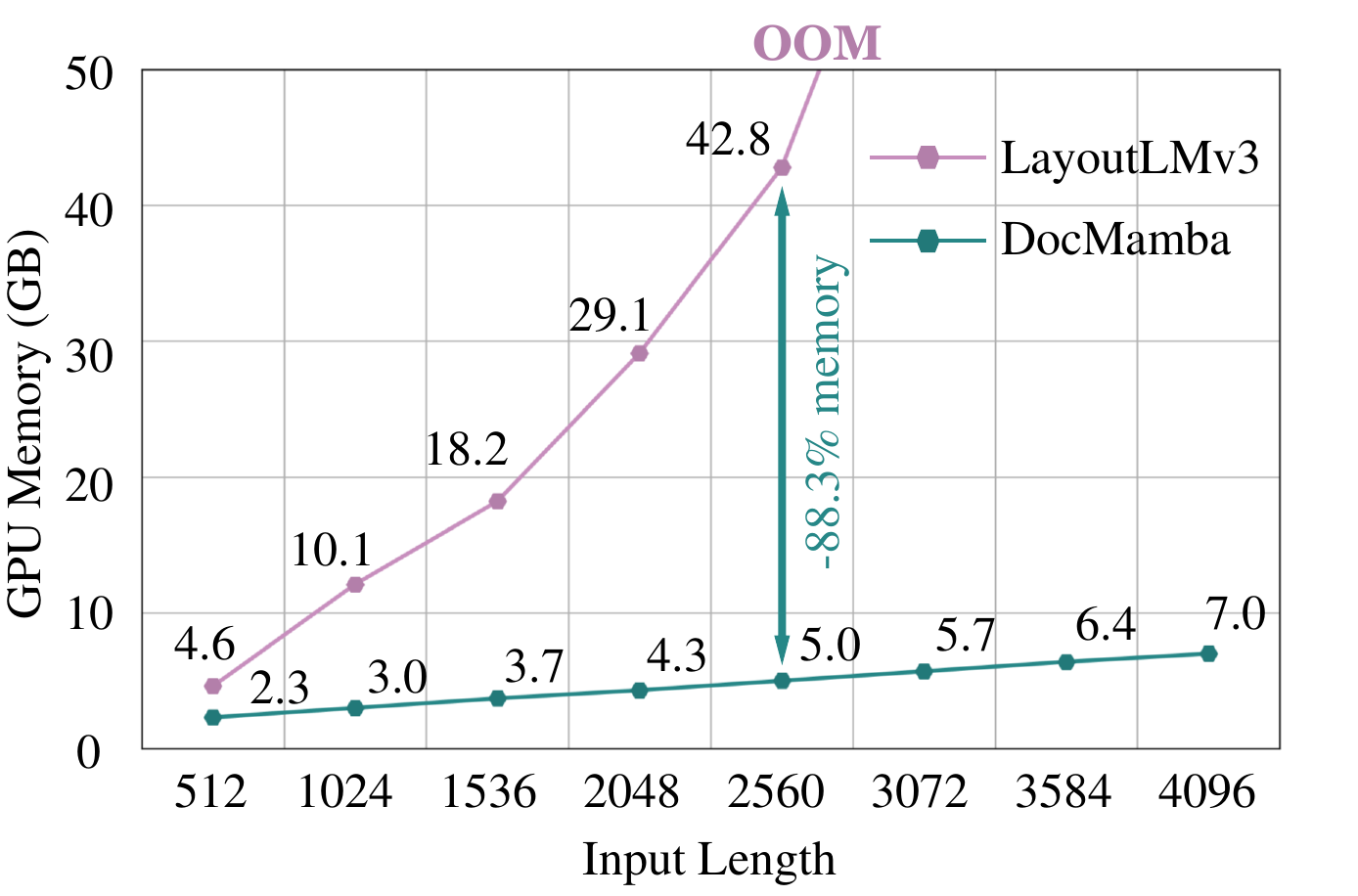} % Reduce the figure size so that it is slightly narrower than the column. Don't use precise values for figure width.This setup will avoid overfull boxes.
	\caption{Comparison of GPU memory usage between LayoutLMv3 \cite{layoutlmv3} and DocMamba.}
	\vspace{-15pt}
	\label{img:memory}
\end{figure}

\begin{figure}[t]
	\centering
	\includegraphics[width=0.99\columnwidth]{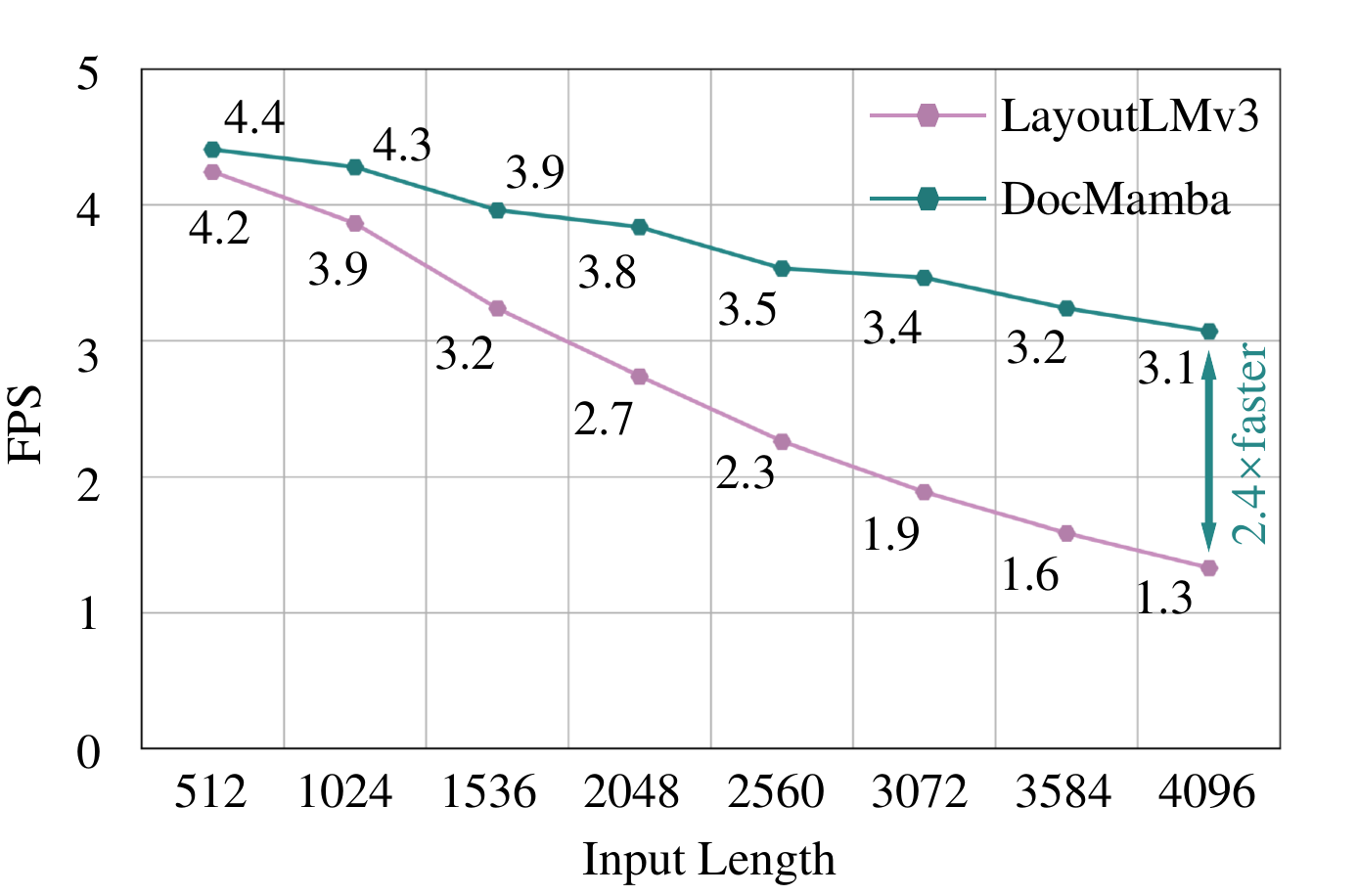} % Reduce the figure size so that it is slightly narrower than the column. Don't use precise values for figure width.This setup will avoid overfull boxes.
	\caption{Comparison of Frames Per Second (FPS) between LayoutLMv3 \cite{layoutlmv3} and DocMamba.}
	\label{img:speed}
\end{figure}

\noindent \textbf{Comparison of Speed and Memory Usage.} Among earlier Transformer-based methods, LayoutLMv3 stands out for its impressive performance and unified structure, making it our primary baseline method. To contrast the speed and memory consumption of DocMamba and LayoutLMv3, we choose HRDoc as the evaluation dataset for semantic unit classification. We use the official implementation of LayoutLMv3 available on Hugging Face \footnote{https://huggingface.co/microsoft/layoutlmv3-base} for benchmarking. Figure \ref{img:memory} illustrates the memory consumption of both models during inference with the batch size set to 16. The memory consumption of LayoutLMv3 escalates rapidly, resulting in an Out-of-Memory situation when the input length reaches 3,072. Conversely, DocMamba's memory consumption grows in a linear manner with the input length, saving 88.3\% of memory when the input length attains 2,560. Figure \ref{img:speed} displays the inference speed of both models during inference. Batch size is set to 8 to avoid Out-of-Memory caused by LayoutLMv3. As the input length increases, the Frames Per Second (FPS) of LayoutLMv3 declines sharply. When the input length reaches 4096, DocMamba's FPS becomes 2.4 times higher than that of LayoutLMv3. These results affirm the efficiency of DocMamba in processing text-dense documents, and also validate DocMamba's linear computational complexity.

\begin{figure}[t]
	\centering
	\includegraphics[width=0.99\columnwidth]{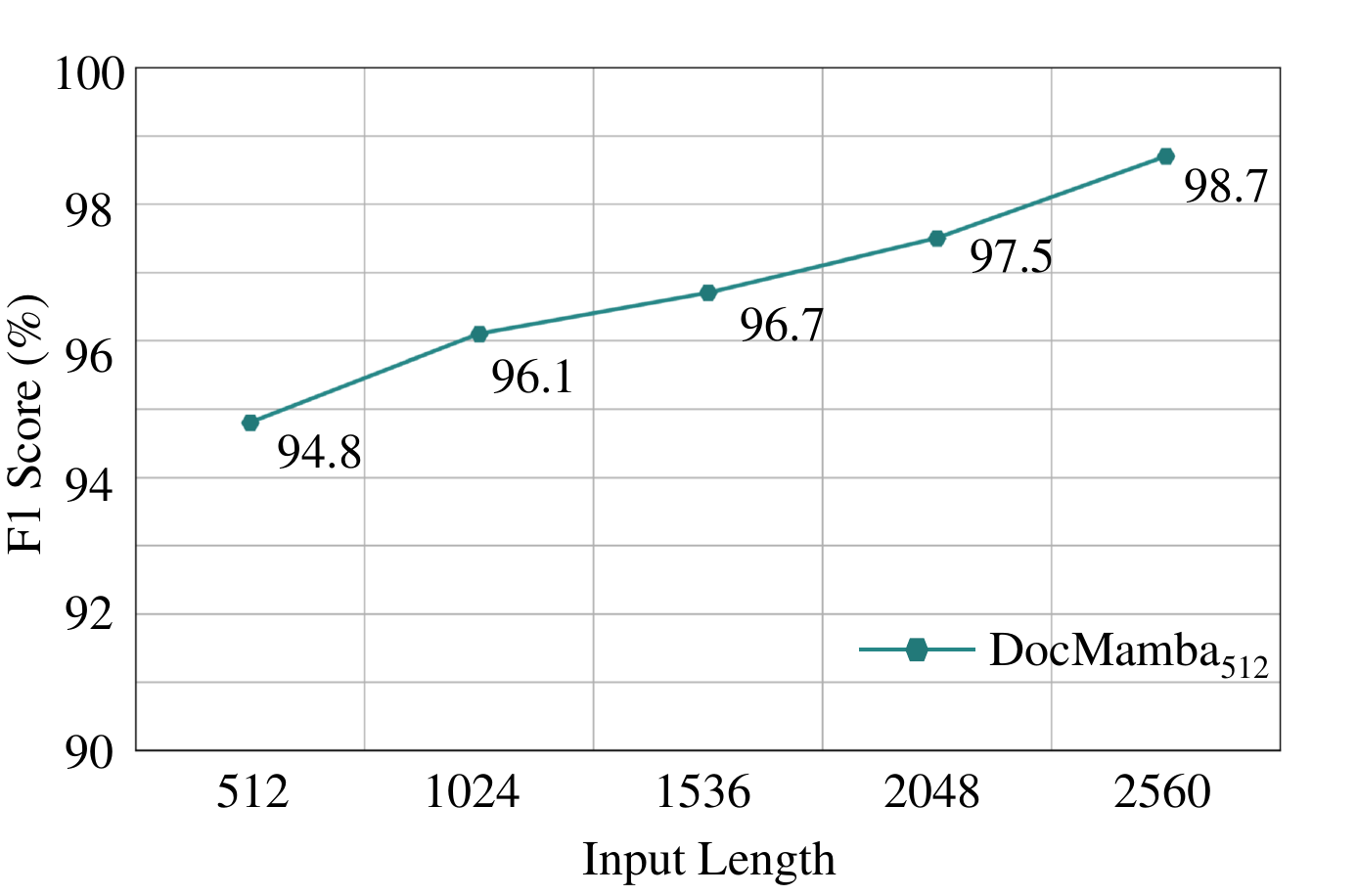} % Reduce the figure size so that it is slightly narrower than the column. Don't use precise values for figure width.This setup will avoid overfull boxes.
	\caption{F1 scores of $\text{DocMamba}_\text{512}$ on HrDoc \cite{hrdoc} with varying input lengths.}
	\vspace{-15pt}
	\label{img:length_extra}
\end{figure}

\noindent \textbf{Comparison of Length Extrapolation.} Transformers  lack an inherent mechanism to consider the order of tokens in a sequence. To address this, many Transformer-based methods in VrDU, such as LayoutLMv3, utilize a learned 1-D position embedding with a prefixed length, which leaves them incapable of length extrapolation. In contrast, SSMs naturally capture sequential and temporal dependencies without a 1-D position embedding requirement, thus endowing DocMamba with length extrapolation potential. We test this feature through the task of semantic unit classification on the HRDoc dataset. We divide document pages based on their length, and select 5 non-overlapping sub-datasets, each spanning a length range of 512. During both pre-training and fine-tuning, we restrict the input length to 512 to obtain the model, $\text{DocMamba}_\text{512}$. The results are illustrated in Figure \ref{img:length_extra}. As the input length increases, the F1 score of DocMamba also sees an upward trend, since the models can leverage longer contexts to yield more precise predictions. This confirms the potential of DocMamba for length extrapolation.

\subsection{Ablation Study}

\noindent \textbf{Impact of Segment-First Bidirectional Scan.} Tokens within documents exhibit complex 2-D spatial layouts. Consequently, we introduce the Segment-First Bidirectional Scan (SFBS) to convert these layouts into 1-D token sequences prior to inputting them into the SSM. To validate the effectiveness of SFBS, we contrast it with the Word-First Bidirectional Scan (WFBS) on FUNSD. Specifically, WFBS utilizes word-level granularity, and organizes tokens directly based on their own Y-axis and X-axis. The order of scanning follows a similar pattern to SFBS. The comparative results are shown in Table \ref{tab:sfbs}. It is clearly evident that the performance of WFBS significantly lags behind SFBS. This can be attributed to SFBS disrupting the sequence of tokens in forms, thereby inhibiting their ability to generate a continuous semantic flow.

\begin{figure}[t]
	\centering
	\includegraphics[width=0.9\columnwidth]{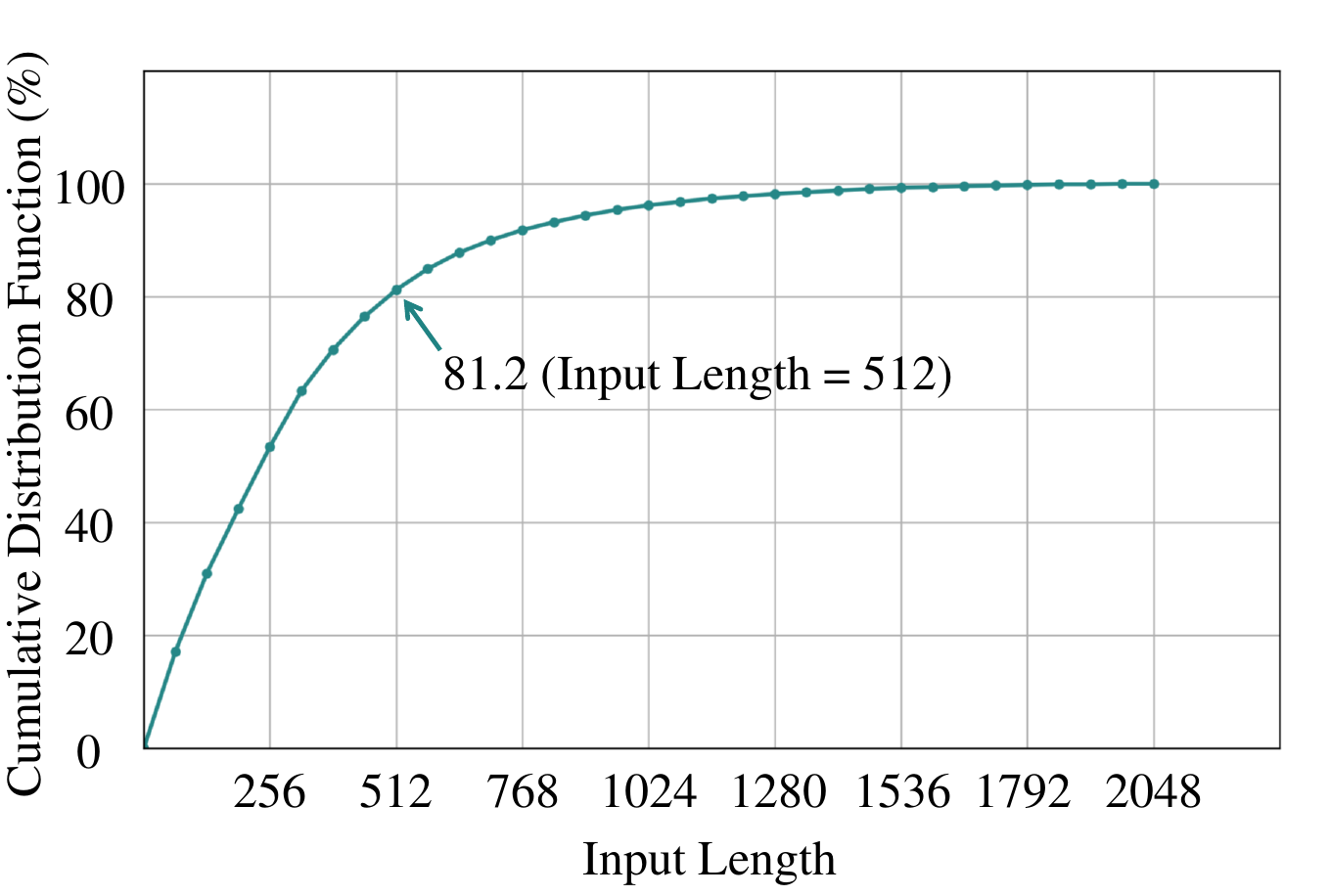} % Reduce the figure size so that it is slightly narrower than the column. Don't use precise values for figure width.This setup will avoid overfull boxes.
	\caption{The cumulative distribution function of input lengths of DocMamba during pre-training.}
	\label{img:input_length}
\end{figure}

\begin{table}[t]
	\centering

	\resizebox{0.72\columnwidth}{!}{%
		\begin{tabular}{llc}
			\toprule
			\bf Scan Strategy & \bf Granularity & \bf FUNSD \\
			\midrule
			SFBS & Segment & {91.7}  \\
			WFBS & Word & {80.8}  \\
			\bottomrule
			\multicolumn{3}{l}{\footnotesize
			}
		\end{tabular}%
	}
	\caption{Ablation study of the Segment-First Bidirectional Scan (SFBS) and Word-First Bidirectional Scan (WFBS). }
	\vspace{-10pt}
	\label{tab:sfbs}
\end{table}

\noindent \textbf{Impact of Input Length in Pre-training.} As introduced in the earlier sections, different from previous Transformer-based methods using a fixed pre-training input length, DocMamba employs a variable input length during pre-training. Figure \ref{img:input_length} showcases the cumulative distribution function of input lengths during pre-training, ranging from 64 to 2,048. To investigate the effect of varying input length, following LayoutLMv3, we limit the input length during pre-training to a maximum of 512 while keeping other settings the same, leading to a new model, $\text{DocMamba}_\text{512}$. The results are presented in Table \ref{tab:docmamba512}. We can make two observations: (1) Increasing the input length is beneficial, as the performance of $\text{DocMamba}_\text{512}$ on the FUNSD and CORD datasets falls short by 0.9 and 0.4 points respectively. (2) Even when the pre-training input length is confined to a maximum of 512, $\text{DocMamba}_\text{512}$ still surpasses LayoutLMv3.

\begin{figure}[t]
	\centering
	\includegraphics[width=0.99\columnwidth]{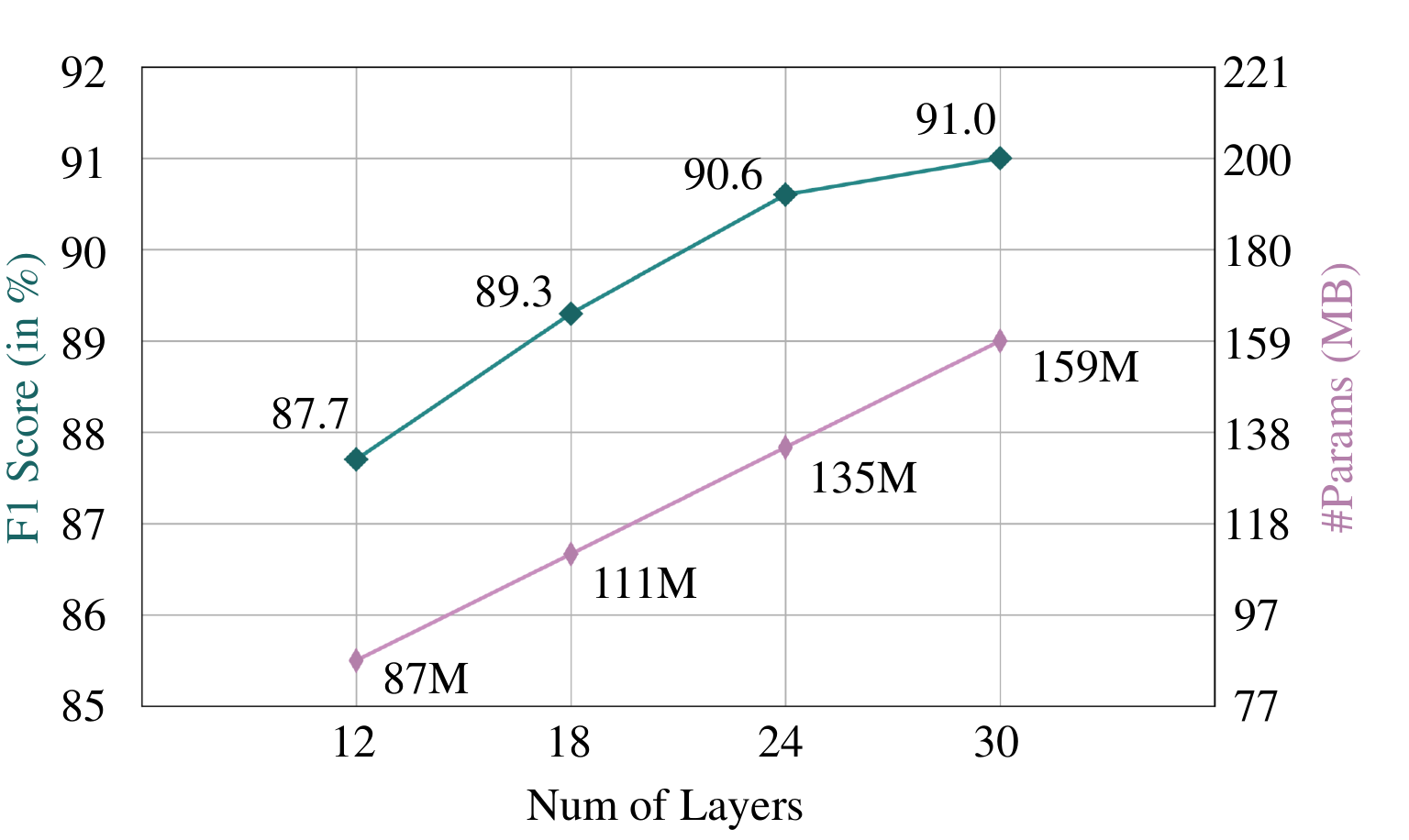} % Reduce the figure size so that it is slightly narrower than the column. Don't use precise values for figure width.This setup will avoid overfull boxes.
	\caption{F1 scores on FUNSD and parameter counts across different layer numbers.}
	\label{img:num_layers}
\end{figure}

\begin{table}[t]
	\centering

	\resizebox{0.78\columnwidth}{!}{%
		\begin{tabular}{lccc}
			\toprule
			\bf Model & \bf FUNSD & \bf CORD  & \bf SROIE \\
			\midrule
			$\text{LayoutLMv3}$ & {90.3} & {96.6} & - \\
			\midrule
			$\text{DocMamba}_\text{512}$ & {90.8} & {96.6} & {96.8} \\
			$\text{DocMamba}$ & {91.7} & {97.0} & {96.8} \\
			\bottomrule
			\multicolumn{4}{l}{\footnotesize
			}
		\end{tabular}%
	}
	\caption{Ablation study of the varying input length. The input length of $\text{DocMamba}_\text{512}$ is limited to 512.}
	\vspace{-10pt}
	\label{tab:docmamba512}
\end{table}

\noindent \textbf{Impact of Number of Layers.} In VrDU, the popularity of Transformer-based models is partially due to their ability to deepen the network by stacking additional layers, facilitating more comprehensive feature learning. Thus, we also explore DocMamba's scalability by adjusting the encoder's layer count to 12, 18, 24, and 30. For experimental efficiency, all models are pre-trained for 10 epochs using the MLM task. The results are presented in Figure \ref{img:num_layers}. A steady increase in the number of parameters can be observed with the rise in layer counts. In addition, DocMamba's F1 score also exhibits a progressive climb, verifying its scalability. This result aligns well with the findings of Mamba in other fields \cite{visionmamba,videomamba}.

%\vspace*{-5pt}

\section{Limitation}
DocMamba's central limitation is its omission of image modality. This decision stems from the observation that DocMamba, employing only text and layout, could already outperform Transformer-based models that incorporate text, layout, and image modalities. This is sufficient to demonstrate the competitive potential of SSM against the Transformer in VrDU. We leave the incorporation of image modality in SSM-based methods to future research in VrDU.

%\vspace*{-5pt}

\section{Conclusion}
In this study, we propose DocMamba, a model based on the SSM that does not rely on the self-attention mechanism. This reduces computational complexity to linear, making it suitable for processing text-dense documents. We also introduce Segment-First Bidirectional Scan, which is used to extract 1-D token sequences from documents. In addition, DocMamba combines text and layout information using a multi-layer bidirectional Mamba encoder. Experiments conducted on publicly available datasets, including FUNSD, CORD, and SROIE, show that DocMamba outperforms previous Transformer-based models, with faster speed  and less memory usage. Further, outcomes on HRDoc validate DocMamba's capacity for length extrapolation. This study highlights the potential of SSM as a powerful tool for understanding visually-rich documents and provides a simple yet effective baseline for future research.

\bibliography{aaai25}

\end{document}